\documentclass[runningheads]{llncs}
\usepackage[T1]{fontenc}
\usepackage{graphicx}
\usepackage{color}
\usepackage{amsmath}
\usepackage{amssymb}
\usepackage{multirow}
\usepackage{booktabs}
\usepackage{cite}
\usepackage[misc]{ifsym}
\usepackage{hyperref}
\hypersetup{colorlinks = True, 
	    linkcolor = red, 
        citecolor = blue}

\begin{document}
\title{
DiffuseReg: Denoising Diffusion Model for Obtaining Deformation Fields in Unsupervised Deformable Image Registration 
}
%
\titlerunning{DiffuseReg}
%
\author{Yongtai Zhuo\inst{1}\orcidID{0009-0005-7400-9280} \and
Yiqing Shen\inst{2}\orcidID{0000-0001-7866-3339}\textsuperscript{(\Letter)}}
%
\authorrunning{Y. Zhuo, Y. Shen}
%
\institute{Shanghai Jiao Tong University, Shanghai, China\\
\email{user\_yuta@sjtu.edu.cn}
\and
Johns Hopkins University, Baltimore, USA\\
\email{yshen92@jhu.edu}
}
\maketitle              
\begin{abstract}
Deformable image registration aims to precisely align medical images from different modalities or times.
Traditional deep learning methods, while effective, often lack interpretability, real-time observability and adjustment capacity during registration inference.
Denoising diffusion models present an alternative by reformulating registration as iterative image denoising.
However, existing diffusion registration approaches do not fully harness capabilities, neglecting the critical sampling phase that enables continuous observability during the inference.
Hence, we introduce DiffuseReg, an innovative diffusion-based method that denoises deformation fields instead of images for improved transparency.
We also propose a novel denoising network upon Swin Transformer, which better integrates moving and fixed images with diffusion time step throughout the denoising process.
Furthermore, we enhance control over the denoising registration process with a novel similarity consistency regularization.
Experiments on ACDC datasets demonstrate DiffuseReg outperforms existing diffusion registration methods by 1.32\% in Dice score.
The sampling process in DiffuseReg enables real-time output observability and adjustment unmatched by previous deep models.
%
The code is available at \url{https://github.com/KUJOYUTA/DiffuseReg}.

\keywords{Denoising Diffusion Model \and Swin Transformer \and Deformable Image Registration \and Consistency Regularization.}
\end{abstract}

\section{Introduction}
\begin{figure*}[t]
\centering
\includegraphics[width=\textwidth]{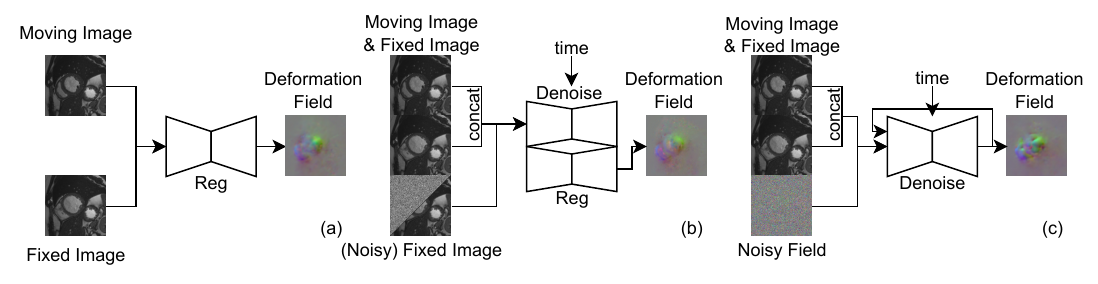}
\caption{
(a) Traditional deep learning registration method, where a moving image is aligned to a fixed image using a learned deformation field generated by a registration network (`Reg').
(b)  Existing approaches claiming to implement diffusion models for registration, shown here without the application of noise during the inference phase, which is signified by the `Noisy' label next to the fixed image.
(c) Our proposed registration with diffusion model. The looped arrow around the denoise network symbolizes the inclusion of a sampling process that iteratively denoises the deformation field.
} \label{fig_Intro}
\end{figure*}

Deformable image registration aims to precisely align anatomical structures across images from different modalities or capture times \cite{abbasi2022medical,zou2022review,chen2021deep}.
As shown in Fig.~\ref{fig_Intro}(a), deep learning-based methods enable end-to-end registration by directly predicting deformation fields from image pairs, reducing inference time \cite{balakrishnan2019voxelmorph,mok2020large,zhao2019unsupervised}.
However, deep learning registration methods lack transparency to observe during the registration process.
They lack flexibility, where they cannot integrate external guidance on deformation characteristics or be observed during inference.
%
%
This limitation means failed registrations typically require costly model retraining rather than efficient adjustment~\cite{balakrishnan2018unsupervised}.

Diffusion models present a promising solution through their inherent conditioning capabilities~\cite{shen2023staindiff,he2023artifact}. 
Moreover, with the denoising sampling process during the inference, applying them to registration could enhance observability and interpretability. 
However, existing diffusion-based approaches like DiffuseMorph \cite{kim2022diffusemorph} and FSDR \cite{qin2023fsdiffreg} primarily manipulate noise in the image space rather than directly in the deformation field, as depicted in Fig.~\ref{fig_Intro}(b).
This divergence from core principle of Denoising Diffusion Probabilistic Model (DDPM)~\cite{ho2020denoising}, which requires noise and outputs to share the spatial domain, leads to the loss of details and suboptimal prediction solely from the lower-dimensional image space.
Another factor that sets those methods apart from DDPM is their omission of the denoising sampling procedure during the registration, which under utilizes the full potential of diffusion models.

Addressing these gaps, we propose DiffuseReg, the first registration diffusion model to generate deformation fields via progressive denoising rather than to manipulate noise in the image space, as shown in Fig.~\ref{fig_Intro}(c).
Our main contributions are three-fold, summarized as follows.
Firstly, we innovatively formulate registration as deformation field denoising in DiffuseReg.
To our knowledge, this represents the first direct application of diffusion models for deformation field generation.
Secondly, we propose a novel Swin Transformer-based denoising network which effectively fuses information from the moving and fixed images with a hierarchical, cross-attention mechanism to guide the denoising process. 
Finally, we improve the quality of the deformation field generation by a novel similarity consistency regularization.

\begin{figure*}[!t]
\centering
\includegraphics[width=0.9\textwidth]{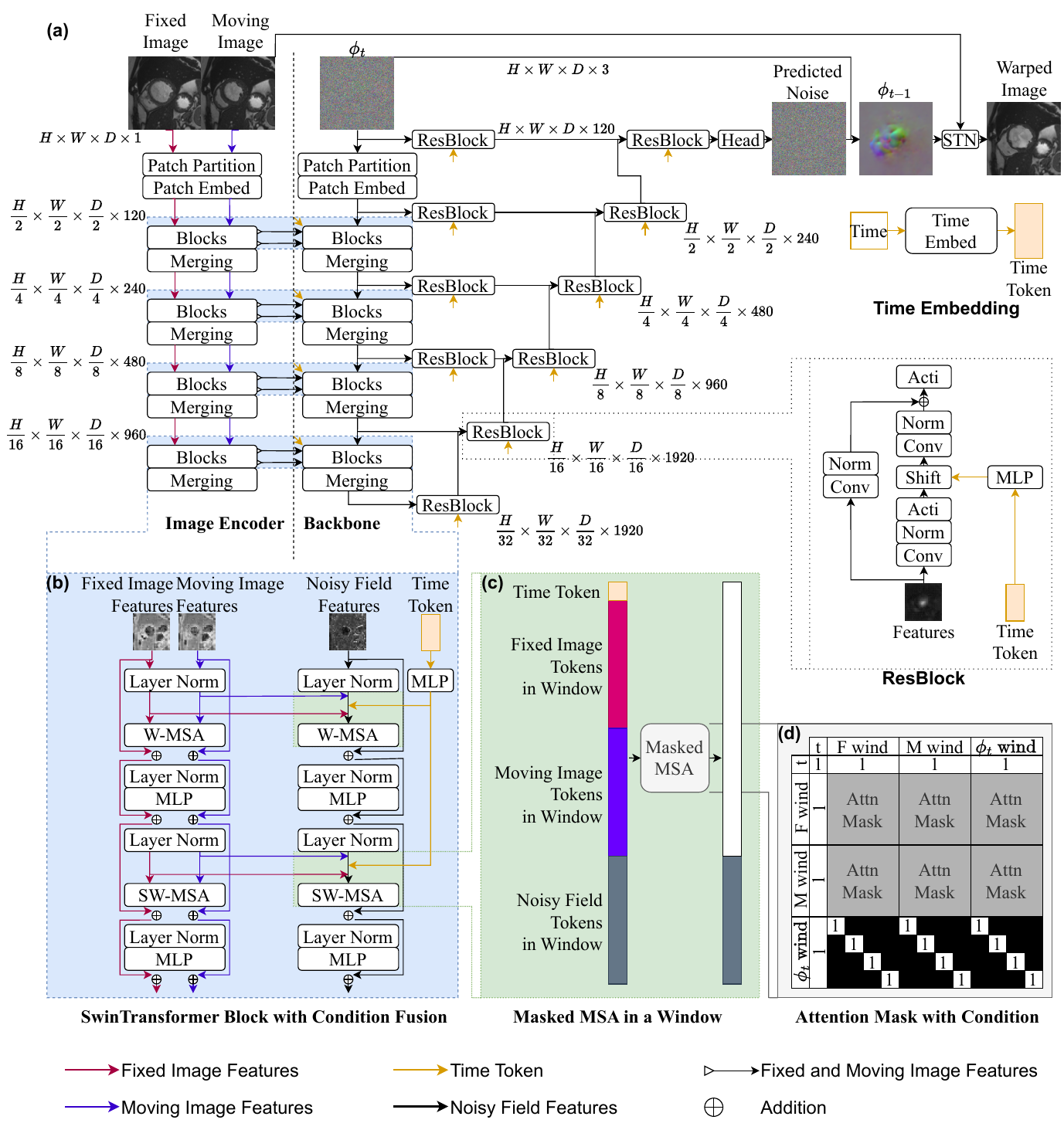}
\caption{
(a) The overall workflow of the denoising network in DiffuseReg. 
(b) Swin Transformer block, showing conditional image feature integration.
(c) The illustration of backbone utilizing conditional information. 
(d) Attention masking in DiffuseReg, with unmasked attention locations indicated by $1$ for selective feature focus.
} \label{fig_Method}
\end{figure*}

\section{Methods}
\subsubsection{Problem Formulation}
The goal of 3D image registration is to compute a deformation field $\phi \in \mathbb{R}^{3\times D\times H \times W}$, representing the spatial mappings along depth, height and width dimensions, that maximizes similarity $\operatorname{sim}(\phi \circ m, f)$ between a warped moving image $\phi \circ m$ and fixed image $f$.
Unlike existing diffusion-based methods \cite{kim2022diffusemorph,qin2023fsdiffreg} that manipulate image noise, we formulate registration using a DDPM to directly predict $\phi$ via progressive denoising. 
Specifically, we initialize $\phi_T$ from the normal distribution $\mathcal{N}(\mathbf{0}, \mathbf{I}_{3\times D\times H \times W})$, conditioned on fixed and moving images. 
The iterative denoising process over timesteps $t$ is framed as
\begin{equation}
\begin{aligned}
\phi_{t-1} = D\left( \phi_t \mid f, m, t \right) \quad \text{where} \quad t=T,\cdots,1,
\end{aligned}
\label{eq1}
\end{equation}
where denoising network $D$ fuses current deformation estimation $\phi_t$ and image contexts $f,m$ to predict the noise.
After convergence at $t=1$, optimal alignment $\phi_0$ is obtained to accurately register moving image $m$ onto fixed image $f$.

\subsubsection{Architecture for Denoising Network}
Successfully registering images requires integrating both local and global contextual information \cite{chen2022transmorph,ha2020semantically}.
Traditional convolutional neural networks (CNNs) like U-Nets \cite{ronneberger2015u} in DDPM often fail to capture broader context, while Transformers \cite{dosovitskiy2020image} struggle with multi-scale visual elements \cite{liu2021swin,he2023artifact}.
Building on Swin Transformer advancements \cite{liu2021swin}, SwinU-Net \cite{cao2022swin} effectively combines local and global insights for medical tasks. 
The enhanced SwinU-NetR \cite{hatamizadeh2021swin} further refines this through CNN ResBlocks in its decoder, enabling smoother output generation. 
However, vanilla SwinU-NetR is limited to unconditional diffusion, contrasting with registration's conditional dependence on fixed and moving images.
To address this, our architecture as depicted in Fig. \ref{fig_Method}(a) consists of two key components, namely an image encoder and backbone.
The encoder leverages Swin Transformer \cite{liu2021swin} as in SwinU-NetR \cite{hatamizadeh2021swin} to separately process the input images and extract feature maps for the backbone.
Mirroring the encoder structure, the backbone encoder fuses these feature maps with the noisy deformation field features and time step embeddings to comprehensively integrate conditions.

\subsubsection{Condition Fusion in Denoising Network}
We propose a novel fusion approach leveraging the Shifted Window Multi-head Self-Attention (S/W-MSA) mechanism from Swin Transformers \cite{liu2021swin}, as shown in Fig. \ref{fig_Method}(b).
Originally used for self-attention within local token windows, we expand S/W-MSA for cross-attention between diverse input sources to enhance registration capability.
The process begins in the image encoder, separately extracting fixed and moving image features.
These are then conveyed to the unified backbone encoder and combined with the current noisy deformation field features and time tokens through S/W-MSA cross-attention.
Specifically, image encoder outputs preceding its S/W-MSA layers are injected into the backbone encoder's corresponding S/W-MSA inputs. 
This allows direct association between all encoder feature types to jointly guide deformation field denoising.

\subsubsection{Masked MSA in a Window} 
Within each window, we concatenate the time token with tokens from the fixed image features, moving image features, and noisy deformation field tokens. 
This combined representation is fed into the masked multi-head self-attention (MSA) operators of the backbone encoder's S/W-MSA layer, as depicted in Fig.~\ref{fig_Method}(c). 
After masked MSA, only the transformed noisy deformation field tokens are retained for subsequent computations while other tokens are discarded:
\begin{equation}
T^{t^{\prime}}, T_i^{F^{\prime}}, T_i^{M^{\prime}}, T_i^{\phi^{\prime}}=\operatorname{masked MSA}\left(\operatorname{concat}\left(T^t, T_i^F, T_i^M, T_i^\phi\right)\right),
\label{eq_MSA}
\end{equation}
where $T^t$ is a time token, $T_i^F$, $T_i^M$ and $T_i^\phi$ are image tokens, moving image image tokens and noisy deformation field tokens within window $i$ respectively. 
Only the transformed deformation tokens $T_i^{\phi^\prime}$ are retained after MSA.
%

\subsubsection{Attention Mask with Conditions}
In masked MSA, attention masks prevent interactions between non-adjacent patches. 
We extend this concept for conditional fusion as shown in Fig.~\ref{fig_Method}(d). 
The attention mask gets repeatedly applied to the attention matrix, enabling sequential fusion between adjacent patch tokens within each window, while keeping full attention between the time token and other tokens.
Specifically, the attention mask is derived from the relative positional encoding between tokens. 
Values of 1 indicate sequential adjacency allowing attention, while 0 masks non-adjacent tokens. 
This selective interaction retains local relevance while allowing the time token to provide extra guidance. 
The deformation tokens freely associate with both local patches and global timestep embeddings through this mechanism.

\subsubsection{Training Objective}
For training, given fixed image $f$, moving image $m$ and the noisy deformation field $\phi_t = \sqrt{\alpha_t} \phi_0 + \sqrt{1-\alpha_t} \epsilon$ following the DDPM formulation, the inputs are fed into the denoising network to predict noise $\epsilon$. 
%
%
The loss function is:
\begin{equation}
\begin{aligned}
& L_{\text {total }}=L_{\text {diffuse }}\left(c, \phi_t, t\right)+\lambda_1 L_{\text {sim }}(f, m \circ \hat{\phi_0})+\lambda_2 L_{\text {reg}}(\hat{\phi_0}), \\
& \text{where } \hat{\phi_0} = \frac{\phi_t-\sqrt{1-\alpha_t} \hat{\epsilon}}{\sqrt{\alpha_t}}.
\end{aligned}
\label{eq5}
\end{equation}
Here $\hat{\epsilon}$ is the predicted noise, $\hat{\phi}_0$ is the predicted deformation field from denoising, and
\begin{equation}
\begin{aligned}
L_{\text {diffuse }}\left(c, \phi_t, t\right)=\mathbb{E}_{\epsilon, \phi_t, t}\left\|D_\theta\left(c, \phi_t, t\right)-\epsilon\right\|_2^2
\end{aligned}
\label{eq6}
\end{equation}
where $\epsilon \sim \mathcal{N}(0, I_{3\times D\times H \times W})$, $c = [f, m]$, and $t$ represents the noise level.

\subsubsection{Consistency Reguarization} Since the $\phi_0$ diffused during training is not ground truth, we introduce additional control terms guiding the denoising process, namely a similarity consistency regulaization $L_\text{sim} = SSIM(f, m \circ \hat{\phi_0})$ and smoothness penalty $L_\text{reg} = \sum\left\|\nabla_\phi\right\|^2$.
Then, same as the inference phase of conditional diffusion, we let the model to progressively sample from Gaussian noise based on conditions as Eq.~\ref{eq1}, ultimately obtaining the deformation field $\hat{\phi_0}$.

\section{Experiment}

\subsubsection{Implementation Details}
We implemented DiffuseReg in PyTorch 1.13. 
The noise schedule was set from $10^{-6}$ to $10^{-2}$ over $T_{train}=2000$ timesteps using linear scaling.
In the training stage, the initialization of the deformation field follows the standard DDPM settings, that is the combination of $\phi_0$ and Gaussian noise.
Pre-trained VTN (ADDD) \cite{zhao2019unsupervised} provided the $\phi_0$ initializations.
We trained for a maximum of 1200 epochs. 
The dissimilarity loss was set as SSIM with kernel size 9, along with $\lambda_1=1$ and $\lambda_2=0.1$ for the consistency terms.
The VTN $\phi_0$ fields were z-score normalized with $\mu=[0.0014,-0.0758,-0.1493]$ and $\sigma=[0.4636,1.1375,1.2221]$ per dimension on the training set.
We used the AdamW optimizer with weight decay $1e-4$, batch size 1 and cosine annealing learning rate scheduling starting from $1e-4$. Training was performed on a single NVIDIA A40 GPU.
In the inference stage, the initialization of the deformation field is pure Gaussian noise.

\subsubsection{Datasets and Metrics}
We utilized the public 3D cardiac MR ACDC dataset \cite{bernard2018deep} for experiments.
It contains 100 and 50 volumetric images pairs for training and testing, depicting the end-diastolic and end-systolic phases along with corresponding segmentation masks.
Following existing conventions \cite{kim2022diffusemorph,qin2023fsdiffreg}, we used the end-diastolic volumes as fixed images and end-systolic as moving for registration.
We utilize dice score to measure the performance and JSD, NJD scores to measure the smoothness of deformation field.

\begin{table}[b!]
\caption{
Quantitative evaluation of image registration on the ACDC dataset.
Dice score measures segmentation overlap for the left ventricle (LV), myocardium (Myo) and right ventricle (RV) regions
Overall average Dice combines all regions, following~\cite{qin2023fsdiffreg}.
Higher Dice indicates better alignment. Lower NJD and JSD indicates better smoothness.
Inference time(s) is also reported.
}\label{tab1-1}
\centering
\begin{tabular}{lccccccc}
\toprule
\multirow{2}{*}{Method} & \multicolumn{4}{c}{Dice(\%)$\uparrow$} & \multirow{2}{*}{JSD$\downarrow$} & 
\multirow{2}{*}{NJD(\%)$\downarrow$} &\multirow{2}{*}{Time(s)} \\
                        & LV & Myo & RV & Overall\\
\midrule
Initial                 &62.32&49.04&66.86& 68.87   & -- & -- & -- \\
DiffuseMorph            &84.24&\textbf{72.54}&78.21& 82.86 &0.53&1.78& 0.3 \\
FSDR                    &83.48&71.40&77.96& 82.30 &0.54& 1.80& \textbf{0.2} \\
Ours                    &\textbf{87.96}&68.62&\textbf{80.09}& \textbf{83.62} & \textbf{0.51} & \textbf{1.74}& 333.2 \\
\bottomrule
\end{tabular}
\end{table}

\subsubsection{Results}
\begin{figure}[t!]
\includegraphics[width=\textwidth]{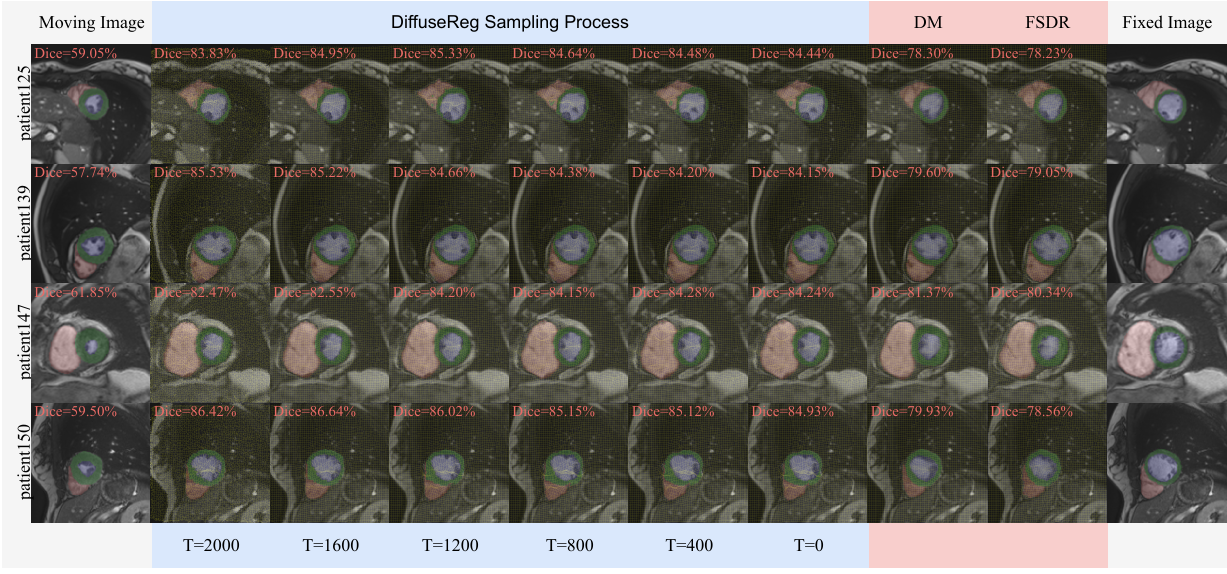}
\caption{
The deformation fields and registered images during sampling process and the comparison with baseline methods. 
The images are annotated with masks of the cardiac tissue. 
The warped images within the blue and red boxes are overlaid with the intricate grid of deformation field.
} \label{fig_Result}
\end{figure}

Table \ref{tab1-1} compares DiffuseReg against existing diffusion-based methods DiffuseMorph \cite{kim2022diffusemorph} and FSDR \cite{qin2023fsdiffreg}.
For the compared methods, we followed the hyperparameters in their paper. 
DiffuseReg achieves the highest overall Dice score, 
and the best JSD and NJD score,
outperforming prior arts in registration accuracy and smoothness while expectedly taking longer due to the sampling process which is lacking in other methods. 
Efficient diffusion sampling method, such as DDIM, can reduce the inference time from 6.5 mins to 0.5 mins without compromising the performance.
Importantly, rather than focusing on the inference speed, our major scope is to provide end users with the opportunity to control the registration process by adding prior information to guide the sampling process via a classifier-guided diffusion scheme, such as control the smoothness of the deformation field, the general direction of deformation and the target organ for registration, where previous end-to-end DL model fails to achieve.
The compared methods generate deformation fields from the image space, which may lead to the loss of details. In contrast, our method directly generates deformation fields in the deformation field space, which helps avoid this issue.
Fig.~\ref{fig_Result} depicts the step-wise outputs from DiffuseReg during sampling, demonstrating progression towards optimal alignment. 
From the image, it can be observed that as the sampling process progresses to 400 steps (t=1600), the noise in the deformation field has been almost eliminated. 
However, further sampling can introduce performance fluctuations attributable to the characteristic additive noise injections of diffusion models. 
Hence the final result may not surpass earlier intermediate states obtained during sampling. 
Therefore, when the generated deformation field is relatively stable, the inference process can be manually stopped to further save the inference time.
This exemplifies the unique tuning capacity provided by DiffuseReg's controllable sampling process. 
Users can observe registrations in real-time to curate ideal outputs unachievable by previous deep approaches lacking such transparency.

\subsubsection{Ablation Study on Denoising Network}
While we utilized SwinU-NetR \cite{hatamizadeh2021swin} as the DiffuseReg denoising network, various backbone choices exist for diffusion models including CNNs and Transformers. We compared several representative networks on the medical image registration task to validate our design in Table \ref{tab2-3}.
Specifically, we adapted U-Net \cite{ho2020denoising}, UViT \cite{bao2023all}, Swin-UNet \cite{cao2022swin} and TransMorph \cite{chen2022transmorph}.
SwinU-NetR achieves superior registration performance over other network variants. 
The comparisons validate it as an optimal backbone balancing local, global and multi-scale reasoning for deformation denoising in DiffuseReg.
We additionally performed ablation experiments validating the impact of the conditional attention masking and time-infused ResBlocks in Table \ref{tab2-3} row 2-4.
First, the conditional attention mask enables selective fusion between adjacent image-deformation tokens and global time tokens.
Removing this masking deteriorates performance by 2.5\% Dice, indicating the importance of focused locality-aware interactions.
Second, the integration of time token into ResBlocks via feature map shifting aids in modeling the dynamic registration process. 
Ablating this time-awareness through removal of shifting operations further reduces performance by 0.5\% Dice.

\subsubsection{Ablation Study on Regularization} 
As shown in rows 1-2 of Table \ref{tab2-3}, inclusion of both $L_{sim}$ and $L_{reg}$, increases the overall Dice score by 2.1\% over the baseline DiffuseReg.
Lack of alignment and smoothness guidance degrades performance, which the proposed consistency terms alleviate.

\begin{table}[t!]
\caption{
Ablation studies on key DiffuseReg components and training strategies. 
`$\phi_0$', `RT', `CM', `CF', `AL' refers to use $\phi_0$ in training phase, Resblocks with time, conditional attention mask, condition fusion method, and additional regularzation loss terms respectively.
}\label{tab2-3}
\centering
\begin{tabular}{cccccccc}
\toprule
$\phi_0$ & RT & CM & AL & CF & backbone & Dice(\%) & NJD(\%) \\
\midrule
$\sqrt{ }$ & $\sqrt{ }$ & $\sqrt{ }$ & $\sqrt{ }$ & Ours & SwinUnetR & 83.6 & 1.74 \\
$\sqrt{ }$ & $\sqrt{ }$ & $\sqrt{ }$ &            & Ours & SwinUnetR & 81.5 & 0.09 \\
$\sqrt{ }$ & $\sqrt{ }$ &            &            & Ours & SwinUnetR & 79.0 & 0.03 \\
$\sqrt{ }$ &            &            &            & Ours & SwinUnetR & 78.5 & 0.04 \\
$\sqrt{ }$ & $\sqrt{ }$ & $\sqrt{ }$ & $\sqrt{ }$ & UViT & SwinUnetR & 71.4 & 0.18 \\
           & $\sqrt{ }$ & $\sqrt{ }$ & $\sqrt{ }$ & Ours & SwinUnetR & 78.0 & 4.31 \\
$\sqrt{ }$ & $\sqrt{ }$ &            & $\sqrt{ }$ & Unet & Unet      & 72.4 & 2.01 \\
$\sqrt{ }$ &            &            & $\sqrt{ }$ & UViT & UViT      & 69.9 & 2.88 \\
$\sqrt{ }$ &            & $\sqrt{ }$ & $\sqrt{ }$ & Ours & Swin-Unet & 77.9 & 2.91 \\
$\sqrt{ }$ &            & $\sqrt{ }$ & $\sqrt{ }$ & Ours & Transmorph& 63.9 & 39.1 \\
\bottomrule
\end{tabular}
\end{table}

\subsubsection{Ablation Study on Fusion Scheme} 
The layer-wise fusion of multi-modal contexts is integral yet complex within DiffuseReg. 
We explored an alternative global concatenation-based approach, fusing information only at the input-level rather than intermediate layers.
Specifically, we first separately embedded the fixed image, moving image and noisy deformation field into patch tokens. 
These tokens were concatenated with time embeddings and passed through a 13-layer UViT network to extract joint representations. 
The deformation tokens were then extracted and fed into our backbone for decoding.
As shown in rows 1 and 5 of Table \ref{tab2-3}, this strategy underperforms layer-wise fusion by 12.2\% Dice, indicating suboptimal guidance for registration without the proposed fusion scheme.

\subsubsection{Ablation Study on $\phi_0$} 
We empirically demonstrate that solely using a similarity loss for training, without the $\phi_0$ initialization and denoising loss, degrades deformation prediction performance.
As observed in row 6 of Table \ref{tab2-3}, training without $\phi_0$ lowers dice score by 5.6\%.
We hypothesize the pre-trained VTN model provides beneficial anatomical priors on reasonable deformations.
Additionally, due to the exclusive use of similarity loss, there were numerous instances of folding in the deformation field. 
Notably, $\phi_0$ guides the diffusion trajectory without directly participating in loss computations. 
%

\section{Conclusion}
In this paper, we introduce DiffuseReg, a novel approach to deformable image registration that leverages the diffusion model to iteratively denoise the deformation field. 
Unlike existing methods that manipulate image noise, DiffuseReg directly applies progressive denoising to the deformation field, adhering closely to the principles of DDPM.
By recurrently estimating registration mappings, DiffuseReg confers unprecedented transparency over the inference process. 
Users can intervene, inspect intermediate outputs, and guide outcomes by tuning noise levels. 
This marks a significant shift from conventional one-shot black-box registration paradigms that lack observability.
One future direction is to reduce the number of sampling steps and decrease the model's inference time. 
Another direction is incorporating information from other modalities as additional input conditions can enhance the controllability and flexibility of the model. 

\begin{credits}
\subsubsection{\discintname}
The authors have no competing interests to declare that are
relevant to the content of this article.
\end{credits}

%
\bibliographystyle{splncs04}
\bibliography{Paper-1193.bib}

\end{document}